# Improved Rapid Serial Visual Presentation using Natural Language Processing

David Awad

Georgia Institute of Technology

davidawad@gatech.edu

*Abstract*—Thoth is a tool designed to combine many different types of speed reading technology. It's largest insight is using natural language parsing for more optimal rapid serial visual presentation and more effective reading information.

**1 INTRODUCTION**

There has been a lot of work in the fields of cognitive neuroscience and psychology on speed reading. The primary focus of a lot of this kind of research generally has been to determine what the capabilities are of people to read faster. There is a lot of content to read out there. It can be thought of in a general sense that there is high, medium, and low fidelity content. Generally we are not so interested in speed reading Hemingway. However we are often interested in speed reading, or skimming, large documents that contain a lot of information we don't need, and only some information we do.

We don't always need to get a complete understanding that has all the details when reading a document. Sometimes just "thumbing through" a document is enough to get a general understanding.

**1.1 Some background on Reading**

When reading documents in general we're actually doing something that's quite new for humanity. The notion of written language itself is a very recent development in human history. The first writing was found in Ancient Mesopotamia, Sumeria and Egypt, with the majority of the oldest samples originating around 3100 B.C. (Schmandt-Besserat, 1986) Not a lot of evolution happens in 5000 years. It may not come as a surprise that when we read, what we're actually using are visual facilities that *are not* evolutionarily optimal for this process. Dehaene and Cohen concluded the eye did not evolve for reading. (Dehaene & Cohen, 2011)

The eyes are pretty good at the process of reading, but what they're doing isn't exactly what we think of when we normally discuss reading. Our brains actually sees (and chunks) words we already know as a single picture instead of reading at the granularity of *individual* letters like we would for a new word we might not know as well like lugubrious, licentiousness, or hagiography[1]. (Gelzer et. al, 2015).

Neurons in a small area of the brain retain what the whole word looks like using what could be called a visual dictionary. This area is called the visual word form area (VWFA) and it resides in the left side of the visual cortex. There's another part of the visual cortex on the other side of the visual cortex that recognizes faces.

**1.2 Supporting Research on Reading in the Modern World**

***Students today are reading content digitally.***—*Students* today read much less on traditional sources such as paper, and much more on electronic sources such as computers, tablets, and cell phones (Cartelli, 2012).

***The visual medium when simplified can make reading easier***—*Using* tools like tachistoscopes[2], researchers Mark D. Jackson and James L. McClelland were able to find that you could effectively measure participants ability to read quickly. They found that faster readers were better able to process the information they were seeing more quickly. They also found that if the encoding of visual information was more efficient, that more of that "processing time" could be put

---

[1] Other candidate words to make my point here were words like nikhedonia, alysm, and shivviness.
[2] A tachistoscope is a device that displays an image for a specific amount of time. It can be used to increase recognition speed, to show something too fast to be consciously recognized, or to test which elements of an image are memorable.

to understanding the content and not interpreting the medium. (Jackson, McClelland, 1975)

***Line Lengths and Screens impacts on reading.***—*Lines* with a length of 55 cpl (characters per line) were found to support effective speed and comprehension. (Dyson, et. al, 2001). They also found that fast readers spend less time between moving from one line to the next. They also found that users were not reading the lines while scrolling, resulting in some wasted time.

They also found "a definition of effective reading therefore needs to differentiate between faster reading and more accurate recall of what is read." (Dyson, et. al, 2001). The most *important* lesson from their research was that "Both very short and very long line lengths can slow down reading through disrupting the normal pattern of eye movements."

> "By reading almost twice as fast as normal we increase the volume of material we get through on screen but acquire a less complete account of a document. Faster readers reading at their normal speed, can recall more than slower readers" (Dyson, et. al, 2001)

***Print Size can improve reading speed.***—*Another* study found that in general slower non-dyslexic readers required larger print size to support their maximum reading speed. (O'Brien, 2005)

***Paper is better for comprehension.***—*Reading* on paper is better than reading on screen in terms of reading comprehension. (Yiren, et. al, 2018) Another study on Norwegian school districts found that we should expect a significant impact on reading performance. (Mangen, 2013). Scrolling specifically seems to really impede reading performance.

*Paper and Digital offer similar speeds.*—*Reading* on paper is not significantly different from reading on screen in terms of reading speed. (Yiren, et. al, 2018)

**1.3 RSVP Tools**

There are tools that have been created to attempt to make it easier to solve different aspects of the reading process and make it faster for users. The most common class of these tools are called RSVP, or Rapid Serial Visual Presentation. These RSVP tools generally iterate through a corpus given by the user, and displays each word, one at a time, through a focused textbox.

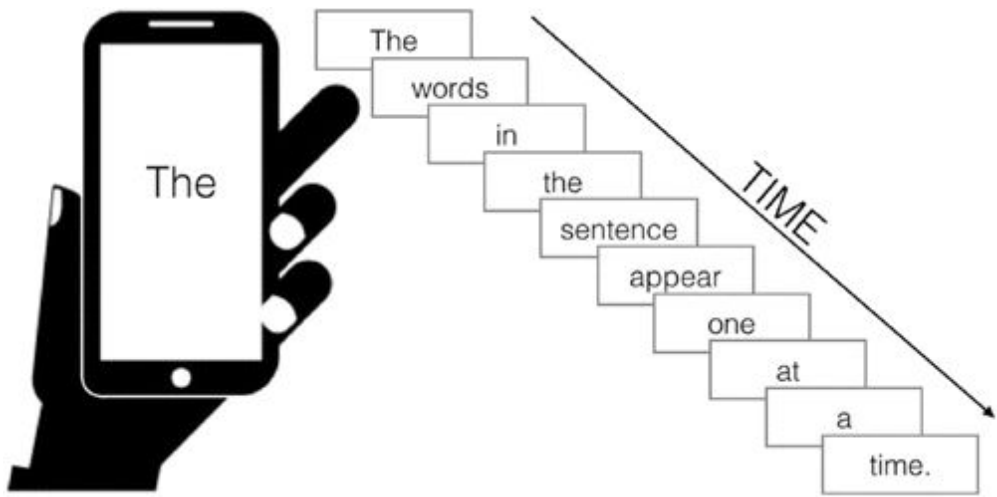

**Figure 1.** Words are displayed through a speed reading device. Source: Rayner, K. sagepub.com

From a technical perspective simply displaying the text alone is not particularly complex, and there have been many tools that simply display words quickly. It's possible to download one of these tools and set the reading speed to 700 words per minute (700 wpm) and each word will be displayed for exactly the same amount of time (roughly 11.6 words per second). This may work for very simple samples of text however just like with normal reading, new words can change the meaning of a sentence, and unfamiliar words require more time for the brain to process and internalize than less familiar words. In addition words of different lengths and complexities have to be dealt with differently. All of this is to say that a lot of conventional tooling around this hasn't made many serious attempts to be more *intelligent* about the problem.

Words in a sentence are not equivalent in meaning, nor are they equivalent length or difficulty to read. Showing words for the same amounts of time is a poor way to go about the problem.

**2 RELATED WORK**

Simple versions of this concept have been built in the form of tools such as [Accelerator](). They are not very technically impressive. They simply display the words faster and one by one.

There are many other applications that simply run similar (if not the exact same) technology such as [Reedy](), [Readsy][3], and [Outread](). The most popular RSVP application by far is Spritz, which seems to be the biggest mover in this space. Spritz does a similar thing to the rest of the pack but has built a good user experience around the reading tool itself.

The biggest issue with a lot of these tools is that they're closed source technology. And it's rare that these businesses actually take off as people usually don't pay for them.

[Zethos]() is an open source version of Spritz that's free and available to anyone. It is one of the few open source tools for speed reading available today, and perhaps the most usable. It's not very well built, but works well for small demonstrations and has gotten some very crucial parts of the user experience right. .

The only other application worth mentioning here is [Beeline](). This company is not using RSVP technology. What they do is illustrate a color gradient along with the paragraph text to solve some of the issues with associating words across line breaks in a large chunk of text. It leverages the fact that visually the colors of the text does impact the association and enables easier reading, especially when reading on something like a [subway car](). It works surprisingly well and is worth being aware of.

Thoth is the first project of it's kind to re-approach the question of RSVP by using machine intelligence instead of the original set-number of milliseconds per word approach. This work sets the stage for applying machine intelligence to the field of RSVP speed reading tools. This is not a novel project in terms of breaking new technical ground; but Thoth is a natural step in a larger narrative of artificial intelligence and natural language processing capabilities being used to improve and disrupt previous methods of doing things. By being open source from the beginning; the stage is set for this work to be improved on and extended by the community.

---

[3] The authors felt it important to note these are real public-facing product names.

It's already been shown that speed reading tools can be more effective in a lot of different situations just by determining the optimal recognition point to more quickly recognize new words through the eyes as images. Where improvement can be made is by using a speed reading tool that can show you the familiar words, or "images", at the faster speed and slowing down at the times where more time needs to be spent for the reader to be able to follow along. This allows a user to get more comprehension while still vastly increasing their reading speed.

## 3 SOLUTION

Thoth is an open source speed reading tool that implements an RSVP speed reader. The biggest differentiators for Thoth are that it's open source, it's ease of use compared to other alternatives, and it's flexibility. Not only does it combine the features of other speed reading tools that exist, it uses natural language processing and artificial intelligence to interpret the difficulty of the text it's being given and customizes it's presentation based on this to the reader.

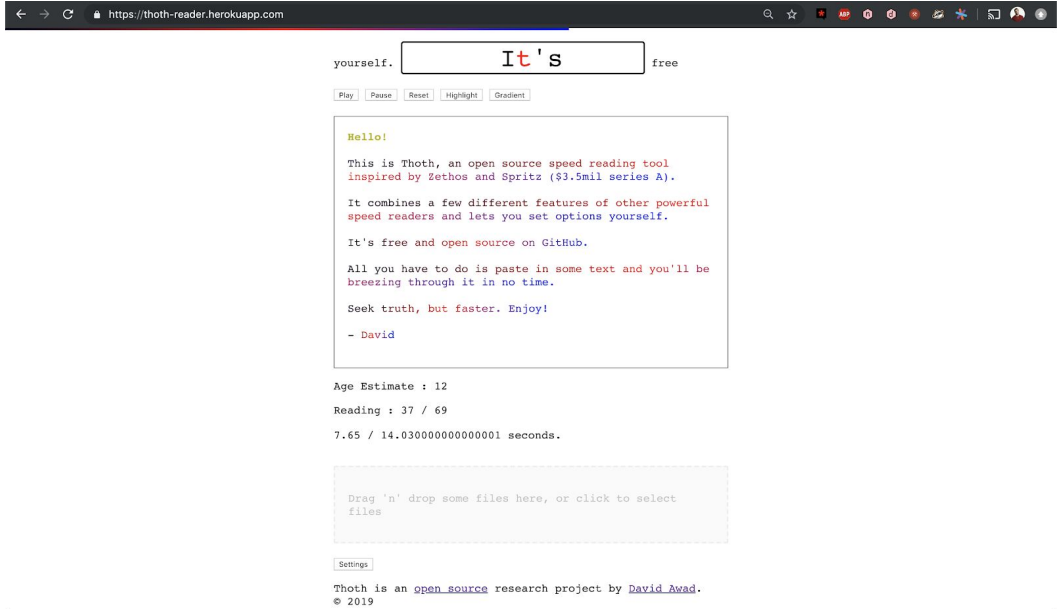

**Figure 2.** A screenshot of the thoth landing page when paused. Source: thoth-reader.

When text or a PDF files is entered into Thoth it parses the text content it receives, and allows the user to apply a color gradient to the text body as well as parse out each word and match a display time to each word. It applies various measurements of the text in order to determine an estimate of the age of the reader that would be required in order to easily understand the text sample given. What's even better is the user can specify their age and the speed will be adjusted for a younger or older reader to be able to move at a more appropriate speed.

## 4 METHODOLOGY

Thoth as a tool is somewhat similar to most speed readers as far as basic functionality. It allows you to input some text and it displays it for you. Where it makes innovations on previous work is around the display time. Thoth uses a few measurements with some different tools to understand more about the text it was given.

It uses measures such as the Automated Readability Index, the Dale–Chall readability formula, the Flesch–Kincaid readability tests, SMOG (Simple Measure of Gobbledygook), Gunning Fog, Spache, and Coleman Liau. With these measurements it can gauge the grade level and age. It also uses Dale–Chall to determine how familiar a word is to a user.

Given that the top 1000 words comprise 80% of the english language it makes sense that our brains would be able to quickly recognize those words and need some more time to recognize other words, Thoth adjusts the display timings to reflect that for each individual word.

**5 RESULTS**

The results have been clear. The tool is freely available to the public and is the only RSVP tool ever created to attempt to use natural language parsing and english language knowledge to better prepare content for rapid presentation.

**6 LIMITATIONS**

It is entirely possible that there are better metrics than Dale–Chall to determine what words are not "familiar" to the average reader and warrant more time. The tool has the ability to use other dictionaries such as Spache or Flesch-Kincaid. In terms of the usefulness of the tool it doesn't *seem* to make a significant different which dictionary is used, as all of them tend to result in a lot of the same words being displayed longer. Dale–Chall turned out to be the most effective for the author.

**7 CONCLUSION**

Thoth is far and away a more usable speed reader than the existing alternatives. With an emphasis on convenience using PDF file parsing and smarter per-word processing, it is much easier for a user to access and load content into it. Thoth has enabled users to read through medium and low fidelity content faster on average while retaining more context and comprehension than they otherwise would have with similar RSVP tools.

# 8 FUTURE WORK

## 8.1 Tunable Parameters

Thoth uses fixed assumptions about how much longer to display a word that is "unfamiliar". There are gradations of complexity in words and it is possible we are losing time by simply scaling the display time of each unfamiliar word by *1.5*.

## 8.2 Other Dictionaries

There is more than one dictionary put out by researchers that claims to easily define the most familiar and most common words. Dale–Chall is only one of those dictionaries that worked nicely for Thoth's use case. There are many others and there is work to be done on which potentially is *most* effective.

## 8.3 User Research

As with many other studies that have appeared in the literature before this, it's quite possible that we could run a study with some potential users, compare Thoth users, with Spritz users, and normal readers and observe how their information compares afterwards.

## 8.4 Speed Writing

The original intent with Thoth was to enable users to more quickly read through text that they wanted to get a general understanding of. The equivalent of a "thumbing through" of the text. One of the next pieces of the project was to attack the problem from the other end. Since there is a dichotomy of *familiar* and *unfamiliar* words within the literature; it stands to reason that you could implement features for speed *writing* rather than just reading.

To put it simply, if you imagine a thesaurus as a mapping of semantically similar words; If you could substitute out difficult words for simpler or familiar words that have the same meaning but are *faster* for the mind to recognize.

That is to say, using natural language processing to explore computationally reducing the english language complexity of text samples given to the speed reader for the brain could make it even faster for a user to read the text. There are some ethical arguments to be made about editing the text on behalf of the user, but this could potentially be a very helpful and effective feature.

**10 APPENDIX**

**Latest Version :**

The latest version of the source code as well as each of the branches for the milestones are freely available on GitHub.com.

**Online Hosted Version :**

The latest stable version is freely available and running on [Heroku](). You can visit it in a browser here: [https://thoth-reader.herokuapp.com]().

**Supporting Open Source Libraries :**

There were numerous javascript libraries that turned out to be really valuable for speeding up my development of Thoth.

- [Compromise js from MIT]()
- The set of linguistic javascript modules from [github/words]()